\documentclass[final]{cvpr}
\usepackage{amsmath,amsfonts,bm}

\def\eqref#1{equation~\ref{#1}}
\def\1{\bm{1}}

\def\vp{{\bm{p}}}
\def\vq{{\bm{q}}}

\def\vx{{\bm{x}}}
\def\vy{{\bm{y}}}

\def\mC{{\bm{C}}}

\def\mW{{\bm{W}}}

\DeclareMathAlphabet{\mathsfit}{\encodingdefault}{\sfdefault}{m}{sl}
\SetMathAlphabet{\mathsfit}{bold}{\encodingdefault}{\sfdefault}{bx}{n}

\def\sX{{\mathbb{X}}}
\def\sY{{\mathbb{Y}}}

\newcommand{\softmax}{\mathrm{softmax}}

\usepackage{times}
\usepackage{epsfig}
\usepackage{graphicx}
\usepackage{amsmath}
\usepackage{amssymb}
\usepackage{multirow}
\usepackage{booktabs} 
\usepackage{caption}
\newcommand{\tabincell}[2]{\begin{tabular}{@{}#1@{}}#2\end{tabular}} 
\usepackage{color}
\usepackage[pagebackref=true,breaklinks=true,colorlinks,bookmarks=false]{hyperref}

\def\cvprPaperID{10117} % *** Enter the CVPR Paper ID here
\def\confYear{CVPR 2021}

\begin{document}

%%%%%%%%% TITLE
\title{Refining Pseudo Labels with Clustering Consensus over Generations \\for Unsupervised Object Re-identification}

% \footnote{The first two authors contribute equally to this work.}

\author{Xiao Zhang\textnormal{\textsuperscript{1}{\small{*}}}
\quad Yixiao Ge\textnormal{\textsuperscript{1}\thanks{The first two authors contribute equally.}}  \quad Yu Qiao\textnormal{\textsuperscript{2,}\textsuperscript{3}}\quad  Hongsheng Li\textnormal{\textsuperscript{1,}\textsuperscript{4}}\\
$^1$CUHK-SenseTime Joint Laboratory, The Chinese University of Hong Kong \quad \\
$^2$SIAT-SenseTime Joint Lab, Shenzhen Institutes of Advanced Technology, Chinese Academy of Sciences\\
$^3$Shanghai AI Laboratory  \quad
$^4$School of CST, Xidian University \\
{
\tt\small zhangx9411@gmail.com}
}

\maketitle

%%%%%%%%% ABSTRACT
\begin{abstract}
   Unsupervised object re-identification targets at learning discriminative representations for object retrieval without any annotations. Clustering-based methods \cite{lin2019aBottom,zeng2020hierarchical,ge2020selfpaced} conduct training with the generated pseudo labels and currently dominate this research direction. However, they still suffer from the issue of pseudo label noise. To tackle the challenge, we propose to properly estimate pseudo label similarities between consecutive training generations with clustering consensus and refine pseudo labels with temporally propagated and ensembled pseudo labels. To the best of our knowledge, this is the first attempt to leverage the spirit of temporal ensembling \cite{laine2016temporal} to improve classification with dynamically changing classes over generations. The proposed pseudo label refinery strategy is simple yet effective and can be seamlessly integrated into existing clustering-based unsupervised re-identification methods. With our proposed approach, state-of-the-art method \cite{ge2020selfpaced} can be further boosted with up to \textbf{8.8\%} mAP improvements on the challenging MSMT17 \cite{wei2018person} dataset. The code is released on \url{https://github.com/2han9x1a0release/RLCC}.

\end{abstract}

%%%%%%%%% BODY TEXT

\section{Introduction}

\begin{figure}[t]
   \begin{center}
   \includegraphics[width=1.0\linewidth]{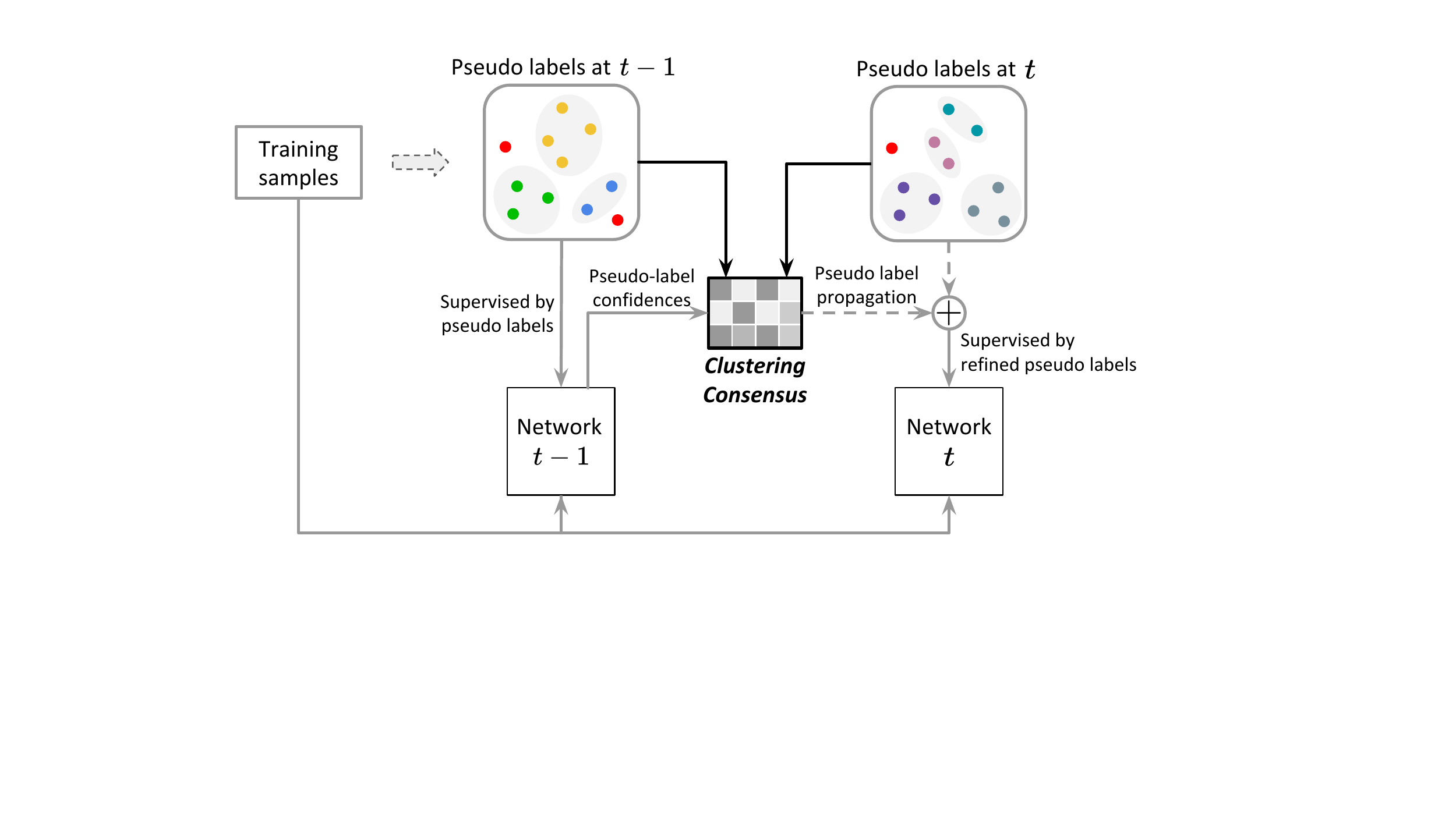}
   \end{center}

      \caption{Illustration of the proposed Refining pseudo Label with Clustering Consensus (RLCC) framework. Hard pseudo labels or soft pseudo-label confidences from the previous generation $t-1$ can be temporally propagated to generation $t$ to effectively refine the pseudo labels at generation $t$ to significantly improve the performance of unsupervised object re-identification.}
   \label{fig:intro}
\end{figure}

Recent years witnessed the remarkable progresses of employing unsupervised representation learning in various downstream visual recognition tasks, such as image classification \cite{caron2018deep,Han2020Automatically,han2019learning,hsu2019multi}, object detection \cite{kim2019self,khodabandeh2019robust,he2019multi,roychowdhury2019automatic}, and object re-identification (re-ID) \cite{lin2019aBottom,lin2020unsupervised,wang2020unsupervised,zeng2020hierarchical,ge2020selfpaced}. Object re-ID aims at retrieving objects of interest in large-scale gallery images given an object's query images.
The task of unsupervised object re-ID further requires learning discriminative representations to properly model inter-/intra-identity affinities without any annotations, which is a more practical setup in real-world applications.

Pseudo-label-based methods with a clustering-based label generation scheme were found effective in state-of-the-art semi-supervised/unsupervised object re-ID approaches \cite{lin2019aBottom,zeng2020hierarchical,ge2020selfpaced,ge2020mutual,zhai2020ad,yang2019selfsimilarity,zhang2019self}. 
An iterative and alternative two-stage pipeline is adopted in each training generation (epoch), \ie, creating pseudo labels and training the network with the generated pseudo labels.
Although multiple attempts on improving the quality of the pseudo labels have been investigated, the training is still substantially hindered by the inevitable label noise, showing noticeable performance gaps compared to the oracle experiments with ground-truth identities \cite{ge2020selfpaced}.
We argue that properly refining the pseudo labels is at the core of further improving unsupervised re-ID algorithms.

To tackle the challenge, we propose a simple yet effective pseudo label refinery strategy following the similar spirit of temporal ensembling \cite{laine2016temporal}, \ie, the pseudo labels from the past generation (epoch) also carry valuable supervision information and can help mitigate the pseudo label noise by smoothing the pseudo label variations.

The temporal ensembling technique has been widely adopted in semi-supervised learning \cite{laine2016temporal,tarvainen2017mean,ge2020mutual} and self-supervised learning \cite{he2019momentum,grill2020bootstrap} tasks. 
It aims at generating more robust supervision signals via aggregating models or predictions with a moving average strategy over previous generations (epochs).
However, it is non-trivial to improve the pseudo labels in unsupervised object re-ID tasks with off-the-shelf label temporal ensembling methods \cite{laine2016temporal,tarvainen2017mean}, since they assume that the class definitions of the recognition tasks remain fixed over training generations.
In contrast, pseudo labels in different training generations for unsupervised re-ID vary much as the pseudo labels are always updated after each generation.

Towards this end, we introduce \textbf{R}efining pseudo \textbf{L}abels with \textbf{C}lustering \textbf{C}onsensus over consecutive training generations (\textbf{RLCC}).
Specifically,
we estimate the pseudo-label similarities over two consecutive generations with an Intersection over Union (IoU) criterion over the sample-label assignments, where a larger value indicates higher consensus between two pseudo classes in two consecutive generations.
To exploit the valuable temporal knowledge encoded by the pseudo labels, we propose to propagate hard or soft pseudo labels from the previous generation to the current generation. The propagation is conducted via a random walk over the pseudo labels, guided by the cross-generation pseudo-label similarities.
Given the temporally propagated labels,
the noisy pseudo labels at the current generation can be properly refined 
via a momentum averaging formulation.
Our proposed refined pseudo labels can be readily integrated into existing clustering-based unsupervised re-ID approaches \cite{lin2019aBottom,zeng2020hierarchical,ge2020selfpaced} with marginal modifications,
\ie replacing the conventional hard pseudo labels with the proposed temporally propagated and ensembled soft pseudo labels.

Our contributions can be summarized into three-fold.
   (1) We introduce to leverage the spirit of temporal ensembling to regularize the noisy pseudo labels in unsupervised object re-ID. Note that existing temporal ensembling techniques \cite{laine2016temporal,tarvainen2017mean} are all designed for close-set classification models, which are not applicable in our task.
   (2) We propose a simple yet effective pseudo label refinery strategy: refining pseudo labels with clustering consensus over training generations (epochs). Our proposed strategy is well compatible with existing pseudo-label-based methods \cite{lin2019aBottom,zeng2020hierarchical,ge2020selfpaced} and leads to further improvements on the already high-performance baseline.
   (3) Our method outperforms state-of-the-arts on multiple benchmarks for unsupervised object re-ID, surpassing state-of-the-art unsupervised method SpCL \cite{ge2020selfpaced} with up to \textbf{8.8\%} mAP improvements.

\section{Related Works}

\paragraph{Unsupervised object re-identification}
requires to learn discriminative representations for object retrieval without any labeled data.
Existing methods \cite{lin2019aBottom,lin2020unsupervised,wang2020unsupervised,zeng2020hierarchical,ge2020selfpaced} mainly focused on training the network with pseudo labels.
A mainstream of methods \cite{lin2019aBottom,zeng2020hierarchical,ge2020selfpaced} adopted clustering algorithms (\textit{e.g.} DBSCAN \cite{ester1996density}) to estimate pseudo labels and were proven effective to achieve satisfactory performance.
BUC \cite{lin2019aBottom} introduced a bottom-up scheme to gradually incorporate more samples in the clusters for training.
HCT \cite{zeng2020hierarchical} encouraged more accurate pseudo labels with a hierarchical clustering algorithm.
More recently, SpCL \cite{ge2020selfpaced} applied a self-paced learning scheme to progressively generate more reliable clusters. 
Although in different ways, they are all devoted to improving the pseudo label quality, which is shown to be the premise of the success of training.

\paragraph{Unsupervised domain adaptation (UDA) for object re-identification}
aims to transfer learned knowledge from the labeled source domain to the unlabeled target domain. 
Existing UDA methods for re-ID can be summarized into two main categories: 
pseudo-label-based methods~\cite{yang2019selfsimilarity, ge2020mutual, ge2020selfpaced, song2018unsupervised, wang2020unsupervised, yu2019unsupervised, zhai2020ad, zhang2019self, zhong2019invariance} and 
domain translation-based methods~\cite{chen2019instance, deng2018image, ge2020structured, wei2018person, yang2020adversarial, xu2019larger},
pseudo-label-based methods are more effective to capture the target-domain distributions.
SSG \cite{yang2019selfsimilarity} estimated multi-scale pseudo labels by leveraging human part features.
MMT~\cite{ge2020mutual} proposed to refine the soft pseudo labels via a mutual learning scheme.
AD-Cluster~\cite{zhai2020ad} refined the clusters with augmented and generated images.
Facing the same problem in unsupervised object re-identification, the major challenge is still on how to provide more reliable pseudo labels and mitigate the noise of the pseudo labels.

\paragraph{Unsupervised representation learning.}
In real-world applications, it is infeasible to annotate a large amount of training data. Therefore, unsupervised representation learning has been widely studied in many computer vision tasks like image classification~\cite{caron2018deep,Han2020Automatically,han2019learning,hsu2019multi,Wang_2020_CVPR, Lin_2020_CVPR, zhang2020discriminability}, image retrieval~\cite{Jang_2020_CVPR}, and object detection~\cite{Hong_2020_CVPR, Ren_2020_CVPR, Chen_2020_CVPR, Wu_2020_CVPR}.
Recently, self-supervised learning methods~\cite{chen2020simple, he2019momentum, hjelm2019learning, oord2018representation, tian2019contrastive, wu2018unsupervised, zhuang2019local} were in favor for unsupervised pre-training tasks, where
a contrastive loss was adopted to learn instance discriminative representations.
However, networks trained by these methods need to be fine-tuned with ground-truth labels on down-stream tasks, which are not applicable in our unsupervised re-identification tasks.

\paragraph{Temporal ensembling} 
was first introduced in semi-supervised learning tasks, forming consensus predictions over training generations.
Laine \etal \cite{laine2016temporal} proposed to use the temporally ensembled predictions as the training targets for unlabeled samples.
Instead of label ensembling, mean-teacher \cite{tarvainen2017mean} proposed to utilize a temporally ensembled model to predict robust supervision signals.
A moving average strategy with momentum was widely used for both model and label aggregating.
The idea of temporal ensembling has also been exploited in self-supervised learning \cite{he2019momentum,grill2020bootstrap}.
Unfortunately, existing label ensembling techniques cannot be directly employed to improve the pseudo label quality for the unsupervised re-identification tasks, since they focused on problems with fixed class definitions over training generations.

\section{Method}

State-of-the-art unsupervised object re-ID algorithms \cite{zeng2020hierarchical,ge2020selfpaced} are based on pseudo labels.
Although a new set of pseudo labels can be generated before each training generation, such pseudo labels have inevitable noise and show large temporal variations over generations, which hinder effective optimization of the re-ID models.

Inspired by temporal ensembling \cite{laine2016temporal}, the pseudo labels and pseudo-label confidences of the training samples from the previous generation can still provide valuable supervision information and also help smooth the pseudo label variations over generations.
The key innovation of our method lies on effectively propagating pseudo labels and confidences from the previous generation to the current one, refining the noisy pseudo labels.
Our proposed label refinery strategy and loss function can be seamlessly integrated into the training of existing approaches without modifying their frameworks or architectures.

\subsection{Revisit of Clustering-based Unsupervised Re-identification}

Pseudo labels were found effective in unsupervised re-ID tasks to provide plausible supervisions on inter-sample affinities for training, where clustering-based label generation schemes dominated recent methods \cite{lin2019aBottom,zeng2020hierarchical,ge2020selfpaced} for achieving state-of-the-art performance.
Given $N$ unlabeled data $\sX$, a two-stage training scheme is alternately adopted in each training generation: (1) generating pseudo labels $\sY$ via clustering the features of the unlabeled training instances, and 
(2) training the network $f_\theta$ with the pseudo cluster labels. Density-based clustering algorithms (\eg, DBSCAN \cite{ester1996density}) would result in outliers, which may serve as distinct instance-level classes as suggested in \cite{ge2020selfpaced}.
Note that we denote a training \textit{epoch} as a \textit{generation} in this paper as 
existing methods iteratively perform the above two-stage scheme with epoch as a cycle.

Naturally, the quality of pseudo labels $\sY$ largely impacts the network capability.
Although pseudo labels can be gradually improved with the proceed of iteratively re-clustering,
we argue that past pseudo labels carry valuable supervision information but were totally ignored by previous methods.

\subsection{Refined Pseudo Labels with Clustering Consensus over Generations}
\label{sec:refine}

We propose to regularize the noisy pseudo labels $\sY^{(t)}$ at the current training generation by exploiting the pseudo labels $\sY^{(t-1)}$ from the past generation.
There are overall $M^{(t)}$ pseudo classes in the current generation and $M^{(t-1)}$ pseudo classes in the previous generation. In general, $M^{(t)} \ne M^{(t-1)}$.

\begin{figure*}[t]
\begin{center}
\includegraphics[width=0.8\linewidth]{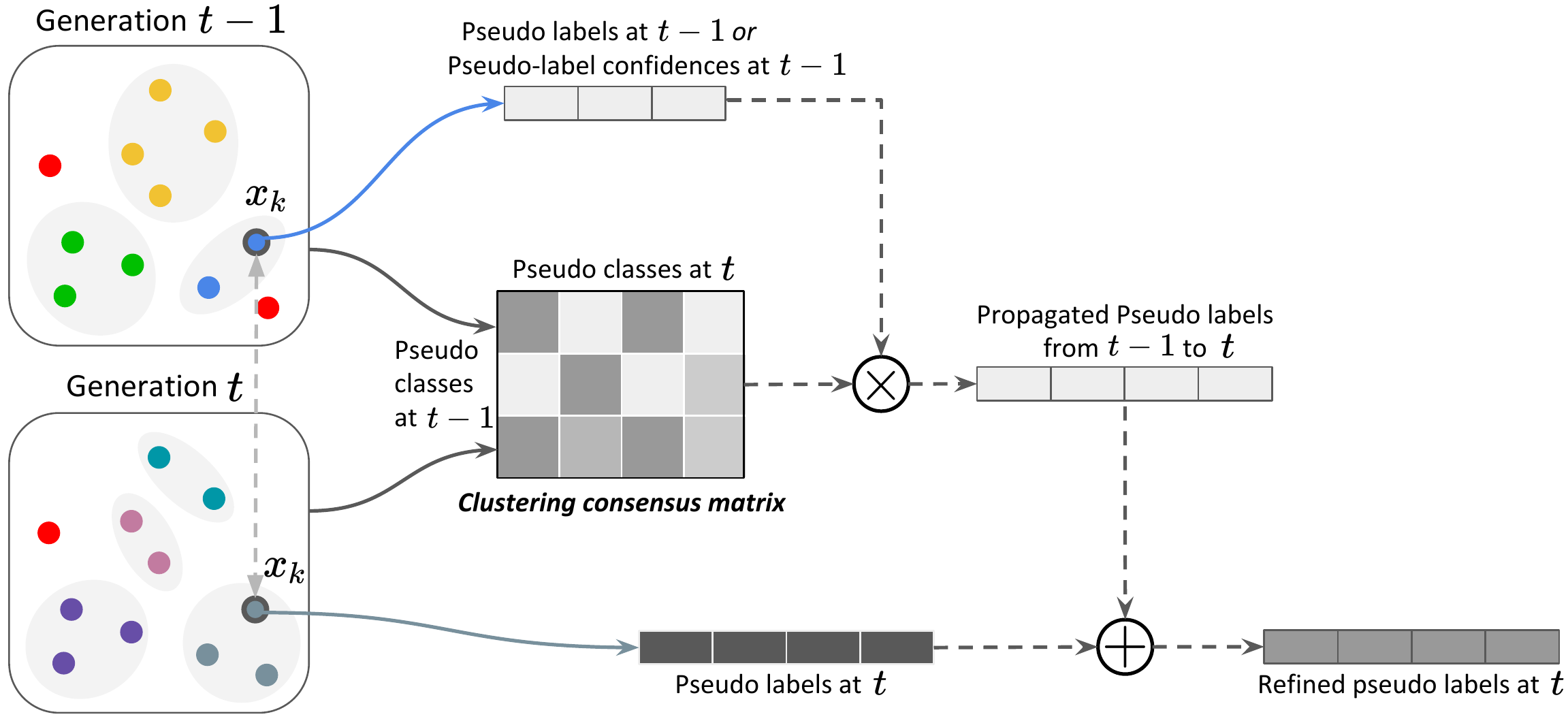}
\end{center}

   \caption{
   Illustration of the overall framework of our proposed Refining pseudo Labels with Clustering Consensus (RLCC) over training generations. The similarities between pseudo labels at generations $t-1$ and $t$ are estimated via their clustering consensus.
   For each sample, either its hard pseudo labels or its pseudo-label confidences at generation $t-1$ can be propagated to generation $t$ according to the cross-generation pseudo-label similarities. The propagated pseudo labels from generation $t-1$ can effectively refine the pseudo labels at generation $t$ to boost the performance of unsupervised object re-identification.}
\label{fig:iou}
\end{figure*}

\paragraph{Clustering consensus over generations.}
Since the label sets over training generations do not overlap, we cannot propagate and aggregate the pseudo labels from the previous generation to the current generation with the off-the-shelf temporal ensembling techniques \cite{laine2016temporal}. 

We therefore propose to first establish the similarities between the pseudo labels of the consecutive two generations, $\sY^{(t-1)}$ and $\sY^{(t)}$, via clustering consensus.
Specifically,
we denote the sample set with a pseudo label of $i$ as $\mathcal{I}^{(t-1)}(i)$ at the previous generation $t-1$, where $i \in [1,M^{(t-1)}]$.
Similarly, samples with a pseudo label of $j$ at the current generation $t$ are denoted as $\mathcal{I}^{(t)}(j)$, where $j \in [1, M^{(t)}]$.
The clustering consensus matrix $\mC \in \mathbf{R}^{M^{(t-1)} \times M^{(t)}}$ is therefore adopted to store Intersection over Union (IoU) criterion between pairs of sample sets from two consecutive generations,
\begin{align} \label{eq:consensus}
    \mC(i,j) = \frac{|\mathcal{I}^{(t-1)}(i) \cap \mathcal{I}^{(t)}(j)|}{|\mathcal{I}^{(t-1)}(i) \cup \mathcal{I}^{(t)}(j)|}\in[0,1],
\end{align}
where $|\cdot|$ counts the number of samples of a set. Intuitively, $\mC({i,j})$ measures the consensus or similarity between the pseudo class $i$ in the previous generation $t-1$ and the pseudo class $j$ in the current generation $t$.
After IoU calculation, we normalize each row of the original consensus matrix $\mC$ to fulfill the constraints that $\sum_j \hat{\mC}(i,j)=1$ for all $j$. The normalization function can be formulated as
\begin{align}
    \hat{\mC}(i,j) = \frac{\mC(i,j)}{\sum_{j=1}^{M^{(t)}} \mC(i,j)},
\end{align}
where the original $\mC$ will be replaced by $\hat{\mC}$ after the normalization, \textit{i.e.} $\mC\leftarrow\hat{\mC}$.

\paragraph{Pseudo label propagation.} 
Given the estimated pseudo label similarities between consecutive training generations, the pseudo label information from generation $t-1$ can be propagated to generation $t$ to refine the current pseudo labels. We investigate propagating two types of pseudo label information from generation $t-1$, (1) hard pseudo labels and (2) soft pseudo-label confidences, for refining pseudo labels $\sY^{(t)}$ at generation $t$.

\noindent {\it (1) Hard pseudo label propagation.} The hard pseudo labels $\sY^{(t-1)}$ encode much information on inter-sample similarities based on the network trained from former generation $t-2$. 
Given an one-hot hard pseudo label $\vy^{(t-1)}_k \in \sY^{(t-1)}$ of the $k$th sample from generation $t-1$, we propose to propagate its previous label to the current generation as
\begin{align} \label{eq:hard}
    \hat{\vy}_k^{(t)} = \mC^\top \vy_k^{(t-1)}, ~~~~\text{ where~~} \vy_k^{(t-1)} \in \mathbf{R}^{M^{(t-1)}}.
\end{align}
The propagated label $\hat{\vy}_k^{(t)} \in \mathbf{R}^{M^{(t)}}$ has the same dimension as the number of pseudo classes at generation $t$. If the ``ground-truth'' pseudo label for sample $k$ is $i$, \ie the $i$th entry of the one-hot vector is $1$ as $\vy_k^{(t-1)}(i) = 1$. The above equation would yield $\hat{\vy}_k^{(t)}(j) = \mC(i,j) \vy_k^{(t-1)}(i)$. In other words, the propagated sample $k$'s pseudo label to the current class $j$ is determined by the cross-generation pseudo-label similarity $\mC(i,j)$ between the pseudo class $i$ at generation $t-1$ and the pseudo class $j$ at generation $t$. In addition, since each row of the $\mC$ matrix sums up to $1$, the propagated labels to the current generation would also sum up to $1$ to ensure they represent a valid confidence vector for supervision, \textit{i.e.} $\sum_j \hat{\vy}_k^{(t)}(j)=1$.
With the proposed propagation scheme, although the two sets of labels have different class definitions (clusters), the pseudo labels can still be successfully propagated across different generations to refine the pseudo labels.

\noindent \textit{(2) Soft pseudo-label confidence propagation.} Although the hard pseudo labels carry some useful information about the feature distributions from network,
the hard assignments of the samples to the pseudo labels make them less robust against label noise. 
Existing temporal ensembling methods \cite{laine2016temporal} have shown that the samples' class confidence vectors from the previous generations can also act as informative training supervisions or regularization for the later generations. 
We take advantages of the key insight and investigate propagating soft pseudo-label confidences from the previous generation $t-1$ to the current generation $t$.
For the $k$th sample $\vx_k \in \sX$, given the network from the previous generation $f_\theta^{(t-1)}$, the sample's classification confidences to pseudo labels at generation $t-1$ can be obtained as
$f_\theta^{(t-1)}(\vx_k) \in \mathbf{R}^{M^{(t-1)}}$, 
where the output dimension of the model matches the number of pseudo labels $M^{(t-1)}$ at generation $t-1$.
Similar to the hard pseudo label propagation, sample $k$'s soft pseudo-label confidences at generation $t-1$ can also be propagated to generation $t$ based on the proposed clustering consensus matrix,
\begin{align} \label{eq:soft}
    \hat{\vy}_k^{(t)} = \mC^\top f_\theta^{(t-1)}(\vx_k), ~~~~\text{ where~~} f_\theta^{(t-1)}(\vx_k) \in \mathbf{R}^{M^{(t-1)}}.
\end{align}
Here $\hat{\vy}_k^{(t)} \in \mathbf{R}^{M^{(t)}}$ denotes the propagated soft pseudo labels from generations $t-1$ to $t$.
The intuition of the propagation is similar to that of hard pseudo label propagation. The soft pseudo-label confidences at generation $t-1$ can be propagated to the current generation $t$ according to the cross-generation pseudo-label similarities by $\mC$. The row-wise normalization property of $\mC$ ensures that the summation of the propagated labels is always up to $1$.
The key difference compared with hard pseudo label propagation is that the model from the previous generation $t-1$ should be kept and used to generate the soft pseudo label for propagation on-the-fly with some extra computational cost.

\paragraph{Pseudo label refinery at generation $t$.}
The propagated pseudo labels $\hat{\vy}^{(t)}_k$ from generation $t-1$ can be integrated into the current pseudo labels ${\vy}^{(t)}_k$ via the momentum averaging formulation
\begin{align} 
    \tilde{\vy}^{(t)}_k = \alpha \cdot {\vy}^{(t)}_k + (1-\alpha)\cdot \hat{\vy}^{(t)}_k,
    \label{eq:temporal}
\end{align}
where $\alpha \in [0,1]$ is a momentum coefficient for ensembling.
By properly estimating the cross-generation pseudo-label similarities in $\mC$, we can refine the original hard pseudo labels with the propagated pseudo labels or pseudo-label confidences from the past generation, \ie pseudo classes that are consistent over the generations would be more confident.
Moreover, the pseudo label variations over consecutive generations can be well smoothed via Eq. (\ref{eq:temporal}), leading to more stable training behavior.

\paragraph{Training objective.}

Our proposed pseudo label refinery strategy is well compatible with existing methods, and can be readily integrated by replacing the hard pseudo labels $\vy^{(t)}_k \in \sY^{(t)}$ with the introduced temporally ensembled soft pseudo labels $\tilde{\vy}^{(t)}_k$ in the training objective, \ie
\begin{align}\label{eq:loss}
    \mathcal{L} = \frac{1}{N}\sum_{k=1}^N \ell_\text{ce} \big(f^{(t)}_\theta(\vx_k), \tilde{\vy}^{(t)}_k\big),
\end{align}
where $\ell_\text{ce}(\vp,\vq)=-\vq\log \vp$ is a cross-entropy loss that uses the refined pseudo labels $\tilde{\vy}_k^{(t)}$ as the training supervisions.

\subsection{Generalization to State-of-the-art SpCL \cite{ge2020selfpaced} Framework}
\label{sec:gen_spcl}

The state-of-the-art unsupervised re-ID method, SpCL \cite{ge2020selfpaced}, has a major difference from conventional clustering-based methods, \ie no classification head is included in $f_\theta$ and $f_\theta(\vx_k)\in\mathbf{R}^L$ indicates the encoded $L$-dimensional feature for $\vx_k$ with normalization.
A non-parametric memory module with dynamic class prototypes $\mW \in \mathbf{R}^{M\times L}$ is adopted in SpCL to replace the classifier to output the class (or pseudo class) logits.

To generalize and integrate our proposed approach into the SpCL framework,
we need to first compute the clustering consensus matrix $\mC$ by Eq. (\ref{eq:consensus}).
The hard pseudo labels or the soft pseudo-label confidences can be propagated from the previous generation $t-1$ to the current generation $t$ guided by $\mC$.
Specifically, the propagation of hard pseudo labels can be computed by Eq. (\ref{eq:hard}).
To propagate soft confidences, as the network $f_\theta$ cannot directly output class confidences, we estimate the confidences with the class prototypes in the memory as $\mW^{(t-1)} f_\theta^{(t-1)}(\vx_k)$.
Eq. (\ref{eq:soft}) then becomes
\begin{align} \label{eq:soft_spcl}
    \hat{\vy}_k^{(t)} = \mC^\top \left(\tau\mW^{(t-1)} f_\theta^{(t-1)}(\vx_k)\right)_\softmax,
\end{align}
where $\mW^{(t-1)}\in\mathbf{R}^{M^{(t-1)}\times L}$ denotes the normalized class prototypes at the generation $t-1$ and $\mW^{(t-1)} f_\theta^{(t-1)}(\vx_k)\in\mathbf{R}^{M^{(t-1)}}$.
$\tau$ is a temperature hyper-parameter for sharpening the class confidences.
Given the propagated label $\hat{\vy}_k^{(t)}$, we can refine the noisy pseudo label ${\vy}_k^{(t)}$ via Eq. (\ref{eq:temporal}).
SpCL adopts a unified contrastive loss as the training objective, which can be treated as a variant of the cross-entropy loss.
To integrate the temporally ensembled soft pseudo labels $\tilde{\vy}^{(t)}_k$ into the unified contrastive loss, 
\begin{align}\label{eq:spcl}
    \mathcal{L} = \frac{1}{N}\sum_{k=1}^N \Big[\ell_\text{ce} \big(\mW^{(t)}f_\theta^{(t)}(\vx_k), \tilde{\vy}^{(t)}_k\big)\Big],
\end{align}
where $\mW^{(t)}\in\mathbf{R}^{M^{(t)}\times L}$.
Although SpCL itself aims at improving pseudo label quality by robustly identifying outliers, our proposed pseudo label refinery strategy with temporal ensembling is well complementary with it, further improving the already strong SpCL (see Sec. \ref{sec:sota}).

% #############################################

\section{Experiments}

\subsection{Datasets and Evaluation Metrics}

\paragraph{Datasets.}
We evaluate our proposed pseudo label refinery strategy on three widely-used person re-ID datasets
and a vehicle re-ID dataset.
\textbf{Market-1501}~\cite{market} 
contains 751 identities for training and 750 identities for testing, captured by 6 cameras.
There are $12,936$ training images, $19,732$ gallery images, and $3,368$ query images.
\textbf{DukeMTMC-reID}~\cite{dukemtmc} 
contains $16,522$ images of $702$ identities for training, and the remaining images out of another $702$ identities for testing. All images are collected from $8$ cameras.
\textbf{MSMT17}~\cite{wei2018person} is a newly released person re-ID dataset with the most images. 
It is composed of $126,411$ person images from $4,101$ identities collected by $15$ cameras.
\textbf{VeRi-776}~\cite{liu2016deep} collects vehicle images in the real-world urban surveillance scenario.
The training set has $575$ vehicles with $37,746$ images and the testing set has $200$ vehicles with $11,579$ images, captured by 20 cameras.

\paragraph{Evaluation metrics.} 
In all the experiments, no ground-truth identities are provided for training.
Mean average precision (mAP) and cumulative matching characteristic (CMC) \cite{market} are adopted to evaluate the methods.
No post-processing technique (\eg re-ranking~\cite{zhong2017re}, multi-query fusion~\cite{market}), is adopted for inference.

\subsection{Implementation Details}\label{ssec:imp_details}

Our proposed pseudo label refinery strategy can be readily integrated into existing clustering-based unsupervised re-ID methods.
To fully verify the effectiveness of our refined pseudo labels, we implement all the experiments based on the state-of-the-art unsupervised re-ID framework SpCL~\cite{ge2020selfpaced}, which is treated as our baseline. DBSCAN \cite{ester1996density} clustering followed by a self-paced strategy is utilized for generating hard pseudo labels after each epoch.
The further improvements based on our approach on the already very strong baseline are convincing to show the superiority of our method. 
We adopt the same settings as used in \cite{ge2020selfpaced} except for the training objective as described in Sec. \ref{sec:gen_spcl}, \ie, we use our refined soft pseudo labels to replace the noisy hard pseudo labels in its unified contrastive loss.
Two hyper-parameters are required in our proposed RLCC strategy, \ie $\alpha$ in Eq. (\ref{eq:temporal}) and $\tau$ in Eq. (\ref{eq:soft_spcl}).
Optimal performances are achieved when adopting the soft pseudo-label confidences propagation with $\alpha=0.9$ and $\tau=30$.

The person images are resized to $256 \times 128$ and the vehicle images are resized to $224 \times 224$. Several data augmentation techniques, such as randomly flipping, erasing~\cite{zhong2017random}, and cropping, are applied to the training samples.
Each mini-batch consists of $64$ images belonging to $16$ pseudo classes are sampled .
An ImageNet~\cite{deng2009imagenet} pre-trained ResNet-50~\cite{he2016deep} is adopted as the backbone for $f_\theta$.
During training, Adam~\cite{kingma2014adam} is adopted to optimize the backbone with a weight decay of $0.0005$. 
The initial learning rate is set to $3.5\times 10^{-4}$ and is decreased to $0.1$ of its previous value every $20$ epochs in the total $50$ epochs.

\subsection{Ablation Studies}\label{ssec:ablation}

In this section, we investigate different designs of our proposed temporally pseudo label refinery, as well as the factors of the hyper-parameters on the Market-1501 dataset \cite{market}. 
In each experiment, all settings are kept the same except for the mentioned one.
Unless otherwise specified, all the experiments in the remaining parts of this section  set $\tau=30$ and $\alpha=0.9$, respectively.

\begin{table}[t]
	\begin{center}
   \begin{tabular}{|l|c|c|c|c|c|}
	\hline
   \multirow{2}{*}{Propagation} & \multirow{2}{*}{$\alpha$} & \multicolumn{4}{c|}{Market-1501~\cite{market}}  \\
   \cline{3-6}
   && mAP & top-1 & top-5 & top-10  \\ 
   \hline \hline
   \multirow{5}{*}{Hard Pseudo Label}&1.00   &  74.1&88.9&95.2&96.8 \\
   &0.95  & 75.4&89.5&95.6&\textbf{97.2} \\
   &0.90 & 75.2 &89.4&95.2&96.9   \\
   &0.85  & \textbf{76.3} & \textbf{90.1} & \textbf{95.8} & \textbf{97.2}   \\
   &0.80 & 75.1&89.0&95.3&96.9   \\
   \hline\hline
   \multirow{5}{*}{\tabincell{l}{Soft Pseudo-label \\ Confidence}}&1.00   &  74.1&88.9&95.2&96.8 \\
    &0.95  & 77.6&\textbf{90.9}&95.8&97.0 \\
   &0.90 & \textbf{77.7} &{90.8}&\textbf{96.3}&\textbf{97.5}   \\
   &0.85  & 76.7 & 90.1 & \textbf{96.3} & 97.0   \\
   &0.80 & 75.1&89.9&95.3&96.7   \\
   \hline

   \end{tabular}
   \end{center}
   \caption{
   Comparison between the hard pseudo label propagation and the soft pseudo-label confidence propagation in our RLCC strategies. The momentum coefficient $\alpha$ varies in $[0.8, 1.0]$ for label ensembling in Eq. (\ref{eq:temporal}).}
   \label{tab:ablation_hard_soft_results}
\end{table}

\subsubsection{Hard \textit{v.s.} Soft Pseudo Label Propagation}

As introduced in Sec. \ref{sec:refine}, we provide two pseudo label propagation strategies, \ie propagating hard pseudo labels via Eq. (\ref{eq:hard}) and propagating soft pseudo-label confidences via Eq. (\ref{eq:soft}).
We evaluate both of the two propagation strategies, as illustrated in Table \ref{tab:ablation_hard_soft_results}.
Since different propagation strategies may have different optimal momentum coefficients for ensembling (Eq. (\ref{eq:temporal})), we conduct experiments with the momentum $\alpha$ varying from $0.80$ to $1.00$.
The results reveal that soft pseudo-label confidences propagation surpasses hard pseudo label propagation consistently under all the hyper-parameter settings, indicating the superiority of propagating soft pseudo-label confidences in the proposed temporally pseudo label ensembling.

\begin{figure}[t]
   \begin{center}
   \includegraphics[width=0.8\linewidth]{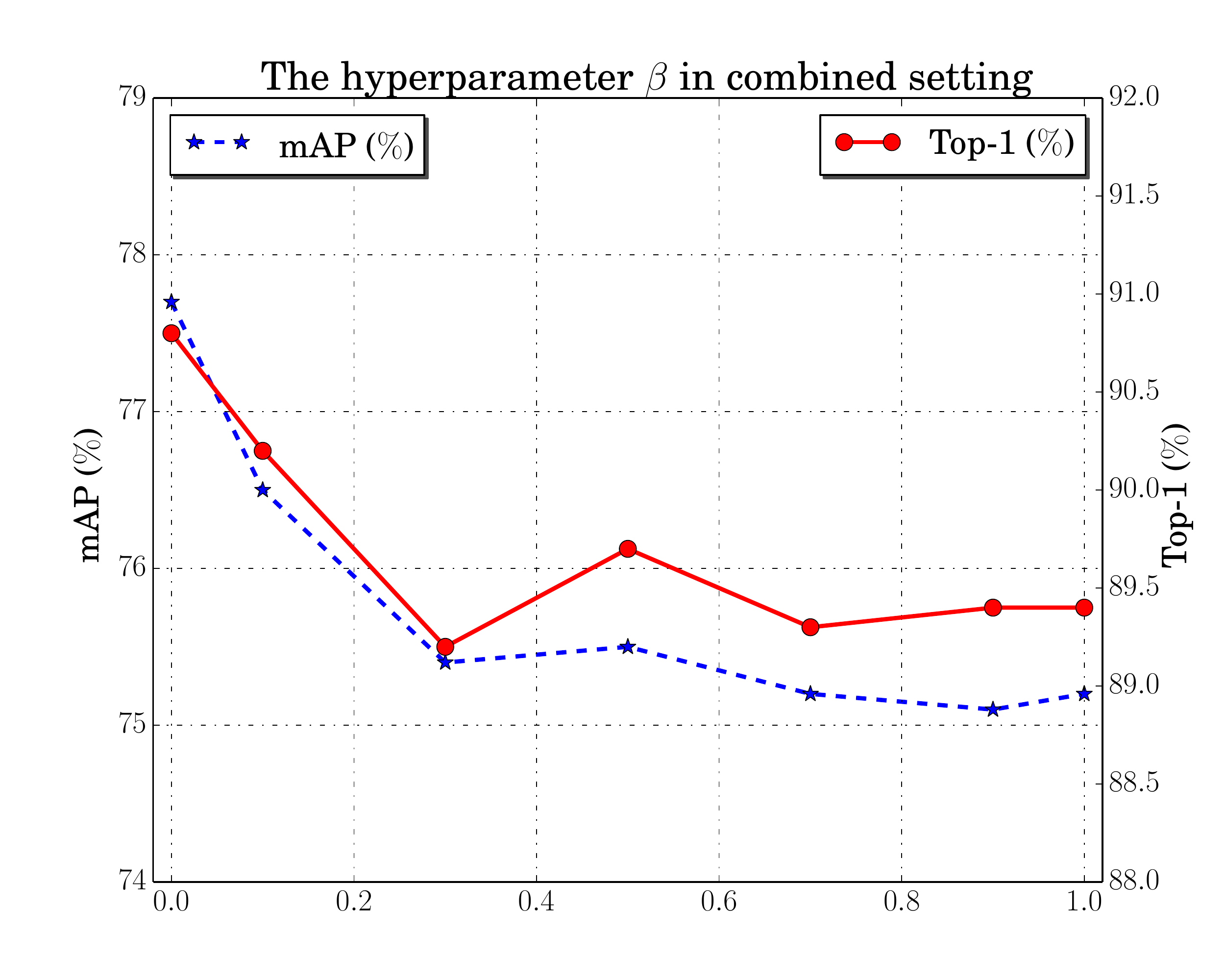}
   \end{center}

    \caption{Performance on Market-1501~\cite{market} when linearly combining the hard pseudo labels and soft pseudo-label confidences for label propagation, as specified in Eq. (\ref{eq:hard_soft}). $\beta$ is used for adjusting the ratio between hard and soft propagation. Results show that soft pseudo-label confidences are more informative than hard pseudo labels.}
   \label{fig:combined}
\end{figure}

To better analyze the correlations between the hard and soft propagation,
we define a general formulation that can reflect the intermediate status between two designs,
\begin{align} \label{eq:hard_soft}
   \hat{\vy}_k^{(t)} = \mC^\top \left( \beta \vy_k^{(t-1)} + (1-\beta) f_\theta^{(t-1)}(\vx_k) \right), 
\end{align}
where 
aggregating hyper-parameter $\beta$ varies from $0$ to $1$.
Eq. (\ref{eq:hard_soft}) is equivalent to the soft propagation Eq. (\ref{eq:soft}) when $\beta=0$, and is equivalent to the hard propagation Eq. (\ref{eq:hard}) when $\beta=1$.
Note that actually, we use Eq. (\ref{eq:soft_spcl}) as the soft propagation to align with the baseline framework of SpCL \cite{ge2020selfpaced}, but for simplicity, we write in the form of Eq. (\ref{eq:soft}) without loss of generality.
By using Eq. (\ref{eq:hard_soft}) for pseudo label propagation and manipulating the hyper-parameter $\beta\in[0,1]$, we achieve the results as shown in Fig.~\ref{fig:combined}.
We observe that the optimal performance is obtained when $\beta=0$, \ie only soft pseudo-label confidences are used for label propagation.
The performance generally decreases as $\beta$ increases.
This phenomenon implies that soft pseudo-label confidences indeed encode more informative supervisions than the hard pseudo labels.

\subsubsection{Class Prototypes for Confidence Estimation}

As introduced in Sec. \ref{sec:gen_spcl},
we use the class prototypes obtained from the non-parametric memory module of SpCL \cite{ge2020selfpaced} to estimate the soft pseudo-class confidences via Eq. (\ref{eq:soft_spcl}).
Specifically, the class prototypes $\mW^{(t-1)}$ from the previous generation $t-1$ is adopted in the estimation of class confidences.
We consider two options of $\mW^{(t-1)}$, one can be cached at the beginning of generation $t-1$ and the other can be computed at the end of generation $t-1$.
Intuitively, the former one carries the most information from the network of generation $t-2$, showing larger temporal variations than the later one, whose supervision information has been mostly tuned at generation $t-1$.
The comparison results can be found in Table~\ref{tab:direct_pred}.
We observe that caching the class prototypes $\mW^{(t-1)}$ at the beginning of generation $t-1$ and then using the kept $\mW^{(t-1)}$ to estimate the pseudo-class confidences for propagation could better refine the pseudo labels, leading to better performance.
The phenomenon indicates that 
properly propagating and ensembling the supervision information is important to refine the pseudo label quality.

\begin{table}[t]
   \begin{center}
   \begin{tabular}{|l|c|c|c|c|c|}
   \hline
   \multirow{2}{*}{$\mW^{(t-1)}$} & \multirow{2}{*}{$\alpha$} &\multicolumn{4}{c|}{Market-1501~\cite{market}}\\
   \cline{3-6}
              &  & mAP  & top-1 & top-5 & top-10 \\
   \hline \hline
   \multirow{3}{*}{\tabincell{l}{Beginning of \\the generation}} & 0.95  & 77.6 & \textbf{90.9} & 95.8 & 97.0    \\
   \cline{2-6}
                     & 0.90   & \textbf{77.7} & 90.8 & \textbf{96.3} & \textbf{97.5}  \\
                     \cline{2-6}
                     & 0.85  & 76.7 & 90.1 & \textbf{96.3} & 97.0    \\
   \hline\hline
   \multirow{3}{*}{\tabincell{l}{End of \\the generation}} & 0.95  & 75.2 & \textbf{89.6} & \textbf{96.3} & 97.5  \\
   \cline{2-6}
                     & 0.90   & \textbf{75.6} & 89.2 & 96.2 & \textbf{97.6}  \\
                     \cline{2-6}
                     & 0.85  & 74.4 & 89.0   & 95.7 & 97.3 \\
   \hline
   \end{tabular}
   \end{center}

   \caption{
   Comparison of class prototypes cached at different training stages. The results exhibit that class prototypes cached at the beginning of generation $t-1$ lead to better pseudo-class confidence estimation for the current generation.}
   \label{tab:direct_pred}
\end{table}

\subsubsection{Hyper-parameter Analysis}

As mentioned in Sec. \ref{ssec:imp_details}, we have two hyper-parameters in our proposed pseudo label refinery strategy.

\paragraph{Momentum $\alpha$ in Eq. (\ref{eq:temporal}).}
\begin{figure}[t]
   \begin{center}
   \includegraphics[width=0.8\linewidth]{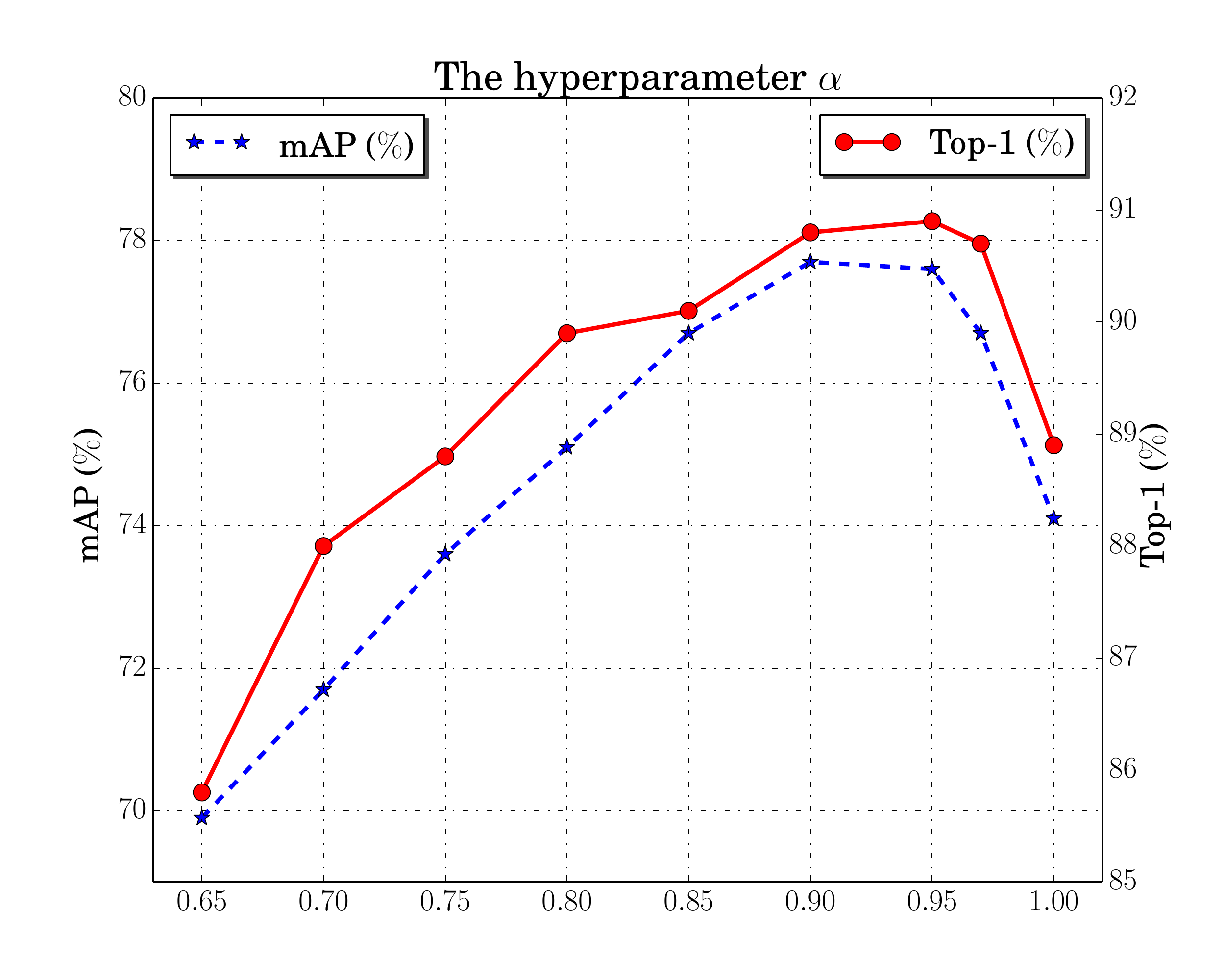}
   \end{center}

      \caption{
      Performances on Market-1501~\cite{market} with different values of the momentum $\alpha$ in Eq. (\ref{eq:temporal}) when fixing $\tau=30$. The model achieves the best mAP=$77.7\%$ when $\alpha=0.9$.}
   \label{fig:chage_alpha}
   \end{figure}
   
As illustrated in Fig.~\ref{fig:chage_alpha}, we analyze the effects of the momentum $\alpha\in[0,1]$ in the ensembling equation. Note the pseudo labels remain the same when $\alpha=1$ and the pseudo labels are totally replaced with the propagated soft labels when $\alpha=0$. Intuitively, the pseudo labels will be smoother with a smaller $\alpha$. However, a too smooth label would raise the issue of information erasing, \eg pseudo labels with uniform distributions contain no information at all. From the experiment results, we find that an optimal performance can be achieved when $\alpha$ is around $0.9$. And the results is robust when $\alpha$ varies from $[0.8,1.0)$, \ie achieving performance gains over the baseline model ($\alpha=1$).

\paragraph{Temperature $\tau$ in Eq. (\ref{eq:soft_spcl}).}
\begin{figure}[t]
\begin{center}
\includegraphics[width=0.8\linewidth]{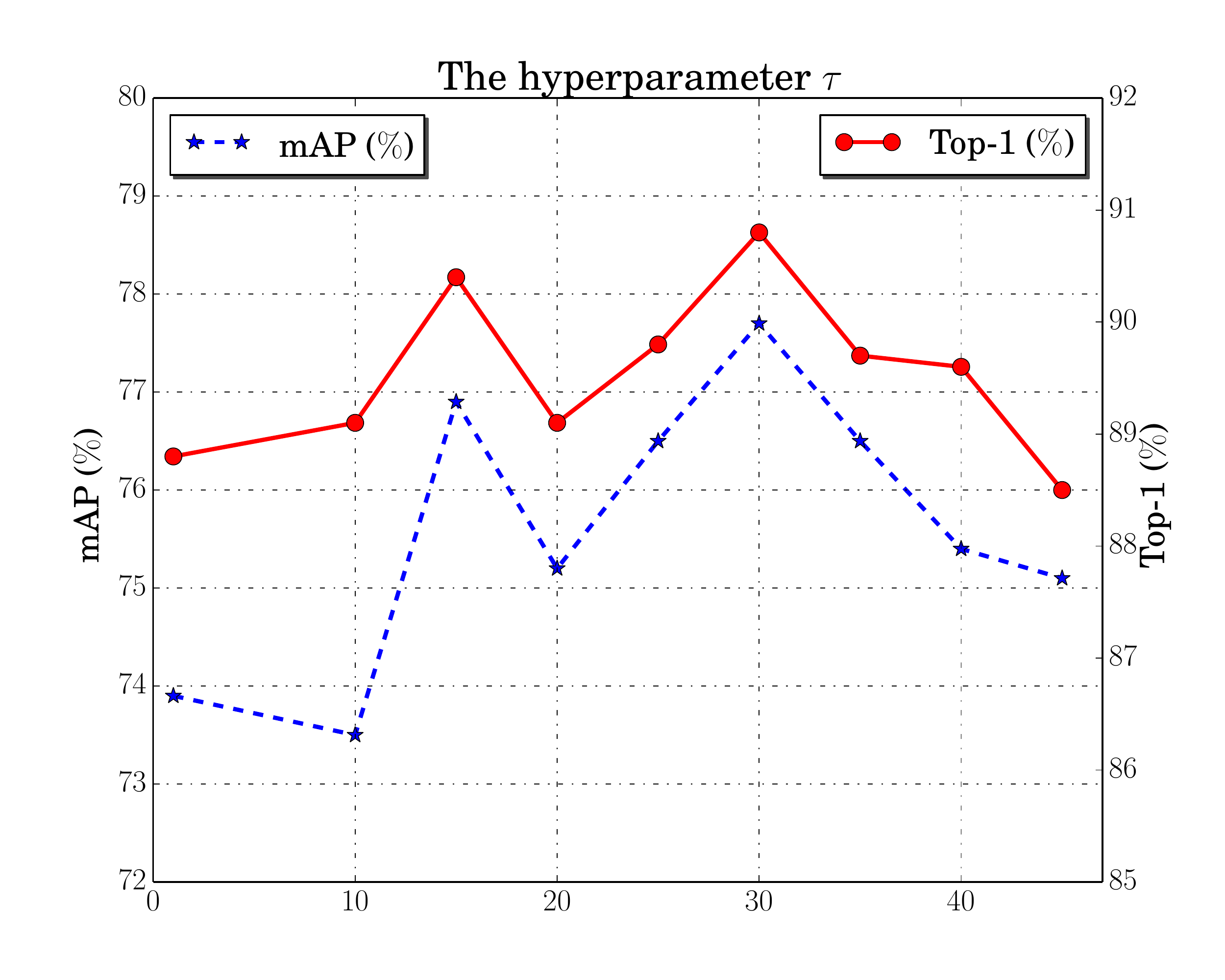}
\end{center}

   \caption{Performances on the unsupervised Market-1501~\cite{market} with different  temperature $\tau$ values in Eq. (\ref{eq:soft_spcl}) when fixing $\alpha=0.9$. The model achieves the best mAP=$77.7\%$ when $\tau=30$.}
\label{fig:chage_s}
\end{figure}

The distribution of original class confidences $\mW^{(t-1)}f_\theta^{(t-1)}(\vx_k)$ estimated by the encoded features and class prototypes tend to be smooth and uniform especially when a large number of pseudo labels exist, since the similarity between the encoded feature and most class prototypes may always be small.
We introduce to adopt a temperature hyper-parameter $\tau$ to sharpen the confidence distribution, leading to more informative soft pseudo-class confidences.
As shown in Fig.~\ref{fig:chage_s}, the optimal performance is obtained when $\tau=30$.
The performances are robust when $\tau$ varies from $15$ to $45$, showing that our method is not sensitive to either hyper-parameter.

\subsection{Comparison with State-of-the-arts}
\label{sec:sota}

We compare our proposed RLCC against state-of-the-art unsupervised re-ID methods on the aforementioned Market-1501~\cite{market}, DukeMTMC-reID~\cite{dukemtmc}, MSMT17~\cite{wei2018person}, and VeRi~\cite{liu2016deep} datasets. The results exhibit that RLCC significantly surpasses all the listed state-of-the-arts in these re-ID evaluation datasets.

\begin{table}[t]
	\begin{center}
   \begin{tabular}{|l|l|c|c|c|c|}
	\hline
   \multicolumn{2}{|c|}{\multirow{2}{*}{Methods}} & \multicolumn{4}{c|}{Market-1501~\cite{market}}  \\
   \cline{3-6}
   \multicolumn{2}{|c|}{} & mAP & top-1 & top-5 & top-10  \\ 
	\hline \hline
   \multicolumn{1}{|l|}{OIM~\cite{xiao2017joint}} & CVPR'17  & 14.0 & 38.0 & 58.0 & 66.3   \\
   \multicolumn{1}{|l|}{BUC~\cite{lin2019aBottom}} & AAAI'19  & 38.3 & 66.2 & 79.6 & 84.5   \\
   \multicolumn{1}{|l|}{SSL~\cite{lin2020unsupervised}} & CVPR'20  & 37.8 & 71.7 & 83.8 & 87.4  \\
   \multicolumn{1}{|l|}{MMCL~\cite{wang2020unsupervised}} & CVPR'20  & 45.5 & 80.3 & 89.4 & 92.3  \\
   \multicolumn{1}{|l|}{HCT~\cite{zeng2020hierarchical}} & CVPR'20  & 56.4 & 80.0 & 91.6 & 95.2   \\
   \multicolumn{1}{|l|}{MMT+~\cite{ge2020mutual}} & ICLR'20  & 74.3 & 88.1 & 96.0 & 97.5 \\
   \multicolumn{1}{|l|}{SpCL~\cite{ge2020selfpaced}} & NeurIPS'20 & 73.1 & 88.1 & 95.1 & 97.0   \\
    \hline 
   \multicolumn{2}{|c|}{\textbf{RLCC}}  & \textbf{77.7} & \textbf{90.8} & \textbf{96.3} & \textbf{97.5}   \\
   \hline
   \end{tabular}
   \end{center}
   \caption{Comparison with state-of-the-art unsupervised person re-ID methods on the Market-1501 dataset~\cite{market}. DBSCAN is applied on both MMT+~\cite{ge2020mutual} and SpCL~\cite{ge2020selfpaced}. Note here MMT+~\cite{ge2020mutual} replaces the average pooling in MMT~\cite{ge2020mutual} by GeM pooling to improve performance and both their results are from OpenUnReID.}
   \label{tab:market_results}
\end{table}

\begin{table}[t]

	\begin{center}
   \begin{tabular}{|l|l|c|c|c|c|}
	\hline
   \multicolumn{2}{|c|}{\multirow{2}{*}{Methods}} & \multicolumn{4}{c|}{DukeMTMC-reID~\cite{dukemtmc}}  \\
   \cline{3-6}
   \multicolumn{2}{|c|}{} & mAP & top-1 & top-5 & top-10  \\ 
	\hline \hline
   \multicolumn{1}{|l|}{OIM~\cite{xiao2017joint}} & CVPR'17  & 14.0 & 38.0 & 58.0 & 66.3   \\
   \multicolumn{1}{|l|}{BUC~\cite{lin2019aBottom}} & AAAI'19  & 27.5 & 47.4 & 62.6 & 68.4   \\
   \multicolumn{1}{|l|}{SSL~\cite{lin2020unsupervised}} & CVPR'20  & 28.6 & 47.4 & 62.6 & 68.4  \\
   \multicolumn{1}{|l|}{MMCL~\cite{wang2020unsupervised}} & CVPR'20  & 40.2 & 65.2 & 75.9 & 80.0  \\
   \multicolumn{1}{|l|}{HCT~\cite{zeng2020hierarchical}} & CVPR'20  & 50.7 & 69.6 & 83.4 & 87.4   \\
   \multicolumn{1}{|l|}{MMT+~\cite{ge2020mutual}} & ICLR'20  & 60.3 & 75.6 & 86.0 & 89.2 \\
   \multicolumn{1}{|l|}{SpCL~\cite{ge2020selfpaced}} & NeurIPS'20 & 65.3 & 81.2 & 90.3 & 92.2   \\
    \hline 
   \multicolumn{2}{|c|}{\textbf{RLCC}}  & \textbf{69.2} & \textbf{83.2} & \textbf{91.6} & \textbf{93.8}   \\
   \hline
   \end{tabular}
   \end{center}
   \caption{Comparison with state-of-the-art unsupervised person re-ID methods on the DukeMTMC-reID dataset~\cite{dukemtmc}.}
   \label{tab:duke_results}
\end{table}

\begin{table}[t]

	\begin{center}
   \begin{tabular}{|l|l|c|c|c|c|}
	\hline
   \multicolumn{2}{|c|}{\multirow{2}{*}{Methods}} & \multicolumn{4}{c|}{MSMT17~\cite{wei2018person}}  \\
   \cline{3-6}
   \multicolumn{2}{|c|}{} & mAP & top-1 & top-5 & top-10  \\ 
	\hline \hline
    \multicolumn{1}{|l|}{MoCo~\cite{he2019momentum}} & CVPR'20 & 1.6 & 4.3 & 9.7 & 13.5 \\
    \multicolumn{1}{|l|}{MMCL~\cite{wang2020unsupervised}} & CVPR'20  & 11.2	& 35.4 & 44.8 & 49.8 \\
    \multicolumn{1}{|l|}{SpCL~\cite{ge2020selfpaced}}  & NeurIPS'20 & 19.1 & 42.3 & 55.6 & 61.2   \\
    \hline 
   \multicolumn{2}{|c|}{\textbf{RLCC}}  & \textbf{27.9} & \textbf{56.5} & \textbf{68.4} & \textbf{73.1}   \\
   \hline
   \end{tabular}
   \end{center}
   \caption{Comparison with state-of-the-art unsupervised person re-ID methods on the MSMT17 dataset~\cite{wei2018person}.}
   \label{tab:msmt17_results}
\end{table}

\paragraph{Unsupervised person re-identiﬁcation.} 
The comparisons with the state-of-the-art algorithms on person re-ID datasets including Market-1501~\cite{market}, DukeMTMC-reID~\cite{dukemtmc} and MSMT17~\cite{wei2018person} are shown in Tables \ref{tab:market_results}-\ref{tab:msmt17_results}, respectively. 
On Market-1501~\cite{market}, we obtain the best performance among all the compared methods with $77.7\%$ mAP. 
Compared to the recent state-of-the-art unsupervised method SpCL~\cite{ge2020selfpaced}, which is also the baseline model of our method, we achieve a noticeable $4.6\%$ mAP improvement. 
On DukeMTMC-reID~\cite{dukemtmc}, RLCC achieves an obvious $3.9\%$ mAP improvement compared to SpCL~\cite{ge2020selfpaced}. 
For the most challenging MSMT17~\cite{wei2018person} benchmark, RLCC achieves an impressive $27.9\%$ mAP, which considerably outperforms state-of-the-art SpCL~\cite{ge2020selfpaced} by an $8.8\%$ mAP improvement.

\begin{table}[t]

	\begin{center}
   \begin{tabular}{|l|l|c|c|c|c|}
	\hline
   \multicolumn{2}{|c|}{\multirow{2}{*}{Methods}} & \multicolumn{4}{c|}{VeRi-776~\cite{liu2016deep}}  \\
   \cline{3-6}
   \multicolumn{2}{|c|}{} & mAP & top-1 & top-5 & top-10  \\ 
	\hline \hline
    \multicolumn{1}{|l|}{MoCo~\cite{he2019momentum}} & CVPR'20 & 9.5 & 24.9 & 40.6 & 51.8 \\
    \multicolumn{1}{|l|}{SpCL~\cite{ge2020selfpaced}}  & NeurIPS'20 & 36.9 & 79.9 & 86.8 & 89.9   \\
    \hline 
   \multicolumn{2}{|c|}{\textbf{RLCC}}  & \textbf{39.6} & \textbf{83.4} & \textbf{88.8} & \textbf{90.9}   \\
   \hline
   \end{tabular}
   \end{center}
   \caption{Comparison with state-of-the-art unsupervised vehicle re-ID methods on the VeRi-776 dataset~\cite{wei2018person}.}
   \label{tab:veri_results}
\end{table}

\paragraph{Unsupervised vehicle re-identiﬁcation.}
We also evaluate our proposed RLCC approach on a popular vehicle re-ID benchmark VeRi-776~\cite{liu2016deep}.
The comparisons with state-of-the-art unsupervised algorithms on VeRi-776~\cite{liu2016deep} are shown in Table~\ref{tab:veri_results}. Compared to SpCL~\cite{ge2020selfpaced}, we achieve a $2.7\%$ mAP improvement. 

The stable performance gains by our RLCC over SpCL~\cite{ge2020selfpaced} reveals that our method can consistently improve the baseline model on various object re-ID task under the unsupervised setting.

\section{Conclusions}

Pseudo label noise is one of the most significant factors that hinder the further improvements of clustering-based unsupervised object re-ID methods \cite{lin2019aBottom,zeng2020hierarchical,ge2020selfpaced}.
To solve this issue, 
we introduce to refine noisy pseudo label with temporally propagated and aggregated soft labels,
which can be readily integrated into existing methods with marginal modifications.
As the label sets vary in different training generations, we propose to estimate clustering consensus to encourage label propagation via a random walk over consecutive generations.
Our success suggests that temporal ensembling with the proposed pseudo-label confidence propagation can effectively mitigate pseudo label noise to achieve higher performance.
Further studies on more properly leveraging the temporal knowledge over more generations are called for.

\noindent\textbf{Acknowledgements.} This work is supported in part by Centre for Perceptual and Interactive Intelligence.

{\small
\bibliographystyle{ieee_fullname}
\bibliography{egbib}
}

\end{document}